\documentclass[sigconf, nonacm]{acmart} %去掉页脚
\AtBeginDocument{%
  }

\usepackage{multirow}
\usepackage{booktabs} % 用于更好的表格线
\usepackage{xltabular} % 在导言区添加

\begin{document}

\title{StockMem: An Event-Reflection Memory Framework for Stock Forecasting}

\author{He Wang}
\email{wanghe@stu.sufe.edu.cn}
%\email{webmaster@marysville-ohio.com}
\affiliation{%
  \institution{Shanghai University of Finance and Economics}
  \city{Shanghai}
  \country{China}
}
% --- 第二作者---
\author{Wenyilin Xiao}
\email{2019212701@live.sufe.edu.cn}
\affiliation{\institution{Shanghai University of Finance and Economics}
\city{Shanghai}
\country{China}}

%通讯作者之一 ---
\author{Songqiao Han}
\email{han.songqiao@sufe.edu.cn}
\affiliation{\institution{Shanghai University of Finance and Economics}
\city{Shanghai}
\country{China}}
%\authornotemark[2]
\authornote{Corresponding authors.}

% 通讯作者之二 ---
\author{Hailiang Huang}
\email{hlhuang@mail.shufe.edu.cn}
\affiliation{\institution{Shanghai University of Finance and Economics}
\city{Shanghai}
\country{China}}
%\authornotemark[2]
\authornotemark[1]
%\authornote{Corresponding author.}


\renewcommand{\shortauthors}{Wang et al.}

\begin{abstract}
 Stock price prediction remains challenging due to market volatility and sensitivity to real-time events. 
 % Traditional numerical stock prediction methods exhibit significant limitations in handling sudden events and shifts in market sentiment. 
  %
   %While large language models (LLMs) offer promise for text-based forecasting, their effectiveness is limited by inadequate memory mechanisms for processing noisy, unstructured financial news. 
  The development of large language models (LLMs) has opened up new possibilities for text-based stock prediction. While LLMs possess powerful text comprehension capabilities, their effective application in complex real-world tasks often relies on specialized memory modules to enable the consistent and stable utilization of relevant historical information. 
  Considerable research has been devoted to constructing such memory architectures.
  However, when confronted with the vast volume of noisy online news data, general-purpose memory architectures face substantial challenges in stock prediction tasks: the text does not contain explicit answers and is filled with noise irrelevant to price movements. The core issue lies in how to autonomously identify and infer the key information and underlying logic that drive stock prices movements from unstructured text.
  To address this challenge, this paper proposes an event-reflection dual-layer memory framework. During the knowledge construction phase, the framework performs in-depth mining of extracted structured events along two dimensions: horizontal event consolidation integrates key daily event content, while longitudinal event tracing tracks the evolution of the same event over time. This process extracts incremental information that reflects market expectation discrepancies, ultimately constructing a temporally-aware structured event knowledge base. Furthermore, by analyzing the dynamic relationship between event sequences and stock price movements, the framework refines stock analysis knowledge to form a reflection knowledge base, thereby achieving systematic and autonomous knowledge construction.
  In the knowledge application phase, the framework involves two steps: retrieval and reasoning. First, the recent event sequence serves as a semantic query to retrieve semantically similar historical scenarios from the memory network, providing precise references for current decision-making. Subsequently, during the reasoning phase, the model synthesizes the recent event sequence, incremental information, and retrieved historical experience to generate reliable predictions of stock price movements.
  Experimental results validate that the proposed framework achieves superior predictive performance over existing memory architectures and exhibits stronger explainability. It can clearly trace the information chain affecting stock prices and analyze the underlying mechanisms of information transmission, endowing the model with professional reasoning capabilities comparable to those of human analysts. This substantially enhances the transparency and trustworthiness of the decision-making process, showing broad application potential in complex financial semantic reasoning tasks.
\end{abstract}


%{Do, Not, Use, This, Code, Put, the, Correct, Terms, for, Your, Paper}


\received{20 February 2007}
\received[revised]{12 March 2009}
\received[accepted]{5 June 2009}

\maketitle

\section{Introduction}
Stock price prediction represents a fundamental and enduring challenge in quantitative finance. The pronounced volatility of stock price sequences, coupled with their acute sensitivity to real-time information, renders traditional numerical prediction methods inadequate for capturing unstructured textual signals such as breaking news and shifts in market sentiment\cite{yaqoob2025predictive,li2022stock,vallarino2025adaptive}. Consequently, these approaches frequently fail to identify the key factors that drive price movements.

%Stock price prediction has long stood as a fundamental challenge in quantitative finance. The highly volatile nature of stock price sequences and their sensitivity to real-time information create significant limitations for traditional numerical prediction methods that rely exclusively on historical prices and technical indicators\cite{yaqoob2025predictive,li2022stock,vallarino2025adaptive}. 
%Such approaches often fail to effectively capture and interpret unstructured textual information like breaking news and market sentiment, frequently missing the critical signals that drive price movements. 

The recent explosion of online textual data, combined with transformative advances in the natural language understanding capabilities of Large Language Models (LLMs), has significantly accelerated research into text-based stock price forecasting. 
%In recent years, the explosive growth of online textual data coupled with transformative advances in large language models' natural language understanding capabilities has driven remarkable progress in text-based stock price prediction research. LLMs can deeply decipher the semantic nuances of financial texts, identifying events, logical relationships, and causal connections within them, thereby establishing a solid foundation for developing more accurate and intelligent prediction models.
Despite this progress, the majority of existing LLM-based studies remain confined to shallow text analysis and immediate inference\cite{vuong2025predicting}. They lack systematic knowledge integration and a long-term memory architecture, which severely limits their ability to adapt to the highly dynamic and complex nature of financial markets. 
These limitations underscore the critical need for external memory mechanisms—models must not only comprehend the semantics of immediate text but also possess the capability to maintain, associate, and retrieve critical information over extended time horizons.
%%股价预测任务的需求
%However, most existing LLM-based studies remain confined to shallow text analysis and immediate inference\cite{vuong2025predicting}, lacking systematic knowledge integration and long-term memory architecture, which makes it difficult to adapt to the highly dynamic and complex nature of financial markets. 
%%

The unique characteristics of financial markets further exacerbate the inadequacy of general-purpose memory architectures. Price fluctuations are seldom triggered by isolated events; instead, they typically emerge from the interplay and sequential evolution of multiple interrelated events over time\cite{yee2025trading}. . 
Furthermore, the paradigm of news-based stock prediction is fundamentally different from traditional question-answering tasks. The prediction target (e.g., "tomorrow's stock price movement") is not explicitly stated within news texts and cannot be directly queried\cite{chen2025stockbench}. This necessitates that models actively construct dynamic market contexts from heterogeneous information flows, identify comparable historical patterns, and perform deep analogical reasoning to achieve a leap from mere "text matching" to genuine "market pattern discovery." 
Compounding these challenges, financial texts are inherently noisy, containing substantial redundant and irrelevant information that can easily obscure signals with genuine predictive value\cite{kamble2025failsafeqa}, making accurate content filtering and matching a paramount concern.

%The unique attribute of financial markets lies in the fact that price fluctuations are typically not triggered by isolated events, but rather stem from the interplay and sequential evolution of multiple interrelated events over time\cite{yee2025trading}. 
%%
%Furthermore, the news-based stock prediction task in this study fundamentally differs from traditional question-answering tasks in its paradigm. The core challenge is that the ultimate prediction target (such as "tomorrow's stock price movement") does not exist as explicit answers within news texts, thus cannot be obtained through direct querying\cite{chen2025stockbench}. 
%This demands the model to actively construct dynamically evolving market contexts from mixed historical and real-time information flows, identify historical patterns comparable to the current state, and perform deep analogical reasoning to achieve a leap from "text matching" to "market pattern discovery." 
%%
%Meanwhile, financial texts generally exhibit high-noise characteristics\cite{kamble2025failsafeqa}, containing substantial redundant and non-critical information that can easily interfere with model judgment. Therefore, how to accurately filter and match content with predictive value becomes a crucial challenge.
%These issues collectively highlight the pressing need for external memory mechanisms—models must not only comprehend the semantics of immediate texts but also possess the capability to maintain, associate, and invoke critical information across extended time spans.

%%现有不足
When applied to stock prediction tasks, general-purpose memory architectures exhibit systematic misalignment across three key dimensions:

$\bullet$ Storage Mechanism: Generic systems typically use isolated documents or facts as storage units. In contrast, financial prediction requires structured event sequences as fundamental units to effectively capture temporal dependencies and joint effects among events\cite{shi2025hybrid}.

$\bullet$ Retrieval Mechanism: Stock prediction relies heavily on analogical reasoning based on historically similar scenarios. Traditional single-document retrieval methods based solely on semantic similarity are ill-suited to support this core requirement.

$\bullet$ Update Mechanism: The recency-based forgetting strategies or simplistic human forgetting curve simulations employed in generic systems often conflict with the financial market's need to preserve long-term effective patterns that retain their relevance.

To address these challenges, this paper proposes a dedicated event memory enhancement framework for financial markets, designed to construct a continuously evolving, professional, and precise repository of financial knowledge and experience for LLMs. The main contributions of this work are as follows:

$\bullet$ We design a structured event-centric memory architecture that converts unstructured market texts into memory units annotated with timestamps, event types, and relevant entities (e.g., companies, industries). This enables explicit modeling of financial event combinations and temporal dependencies, organizing discrete market information into semantically connected structured sequences and laying a solid foundation for subsequent retrieval and reasoning.

$\bullet$ We develop an event sequence similarity-based retrieval mechanism that overcomes the limitations of traditional single-document semantic retrieval. By using the current event sequence as a query, our framework can accurately locate complex historical patterns analogous to the present market context. This provides crucial support for in-depth analogical reasoning, facilitating the transition from "text matching" to "pattern discovery."

$\bullet$ We validate the framework's effectiveness through systematic experiments. Our approach demonstrates superior stock prediction accuracy compared to strong baseline models. Moreover, thanks to the explicit modeling of event sequences, the framework can clearly identify and present the key events influencing stock price trends along with their dynamic evolution paths, significantly enhancing the credibility of predictions and the transparency of the decision-making process.

This research not only advances the application of LLMs in the domain of financial forecasting but also provides a new paradigm for the design of memory mechanisms in specialized domains. By constructing a domain-specific memory system, we effectively address the adaptation problems of general frameworks in professional scenarios, thereby establishing a solid foundation for future related research.

\begin{figure*}[h]
  \centering
\small
\includegraphics[width=\linewidth]{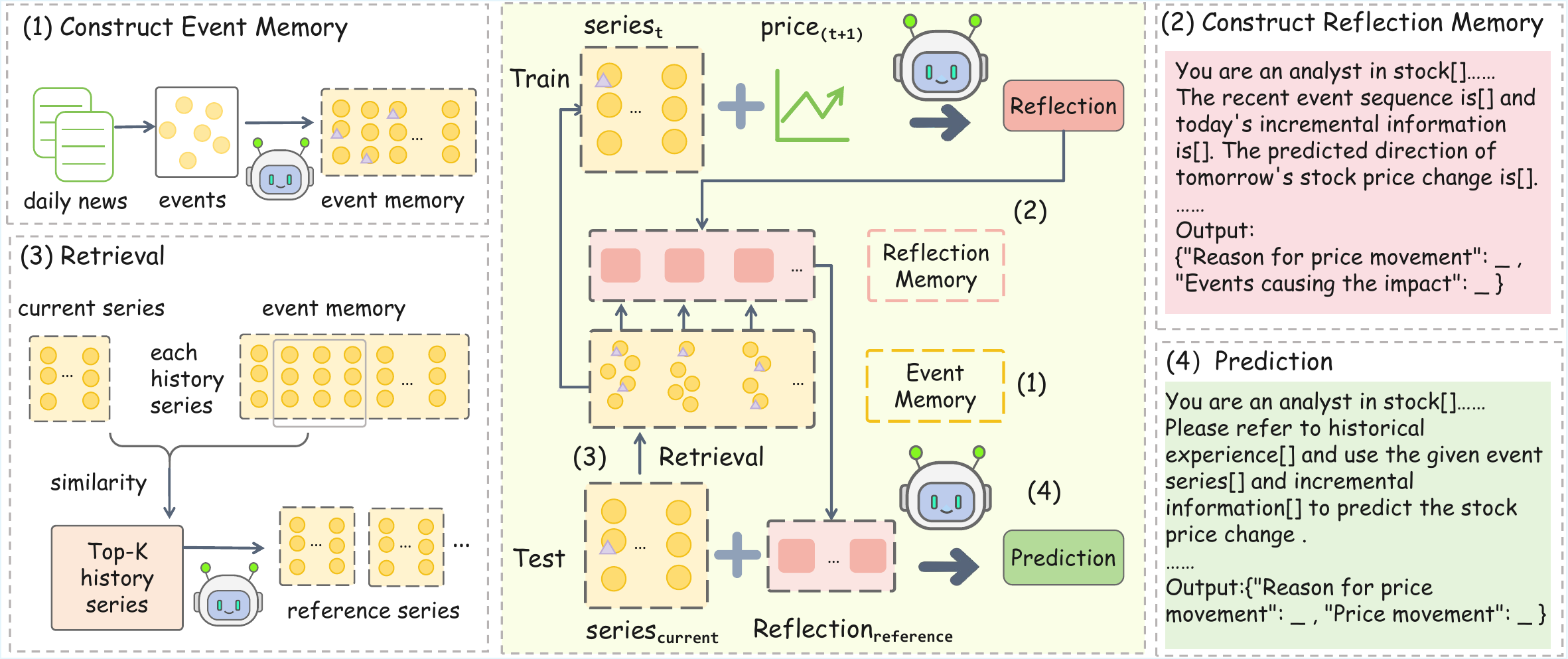}
  \caption{The overall architecture of the StockMem framework.}
  \Description{The overall architecture of the StockMem framework.}
  \label{fig:framework}
\end{figure*}

%\subsection{Relevance }
%This work aims to tackles core challenges in semantic and knowledge systems by introducing a novel framework that constructs and reasons over temporal knowledge bases derived from web-based financial news. 
Our contributions align with the track's scope in three key aspects:
(1)Structured Knowledge Extraction: We convert noisy web news into a temporally-structured event knowledge base, supporting scalable creation of Web-based structured knowledge.(2) Semantic-LLM Synergy: Our dual-layer memory enhances collaboration between structured knowledge and the reasoning ability of LLM.(3)Web Application: We demonstrate practical value in financial prediction, enhancing interpretability for web news-based decision systems.We addresses the essential challenge of deriving dynamic semantic knowledge from web corpora, fitting squarely within the track's focus.

\section{Related Work}
\subsection{Memory Mechanisms for Large Language Models}

Memory mechanisms have become a critical component for enhancing large language models' capacity to handle complex tasks, undergoing an evolutionary path from parametric internal storage to external memory systems. Parametric memory implicitly encodes knowledge within the model's weights, offering advantages in rapid access and persistent retention, yet it inherently suffers from poor interpretability and difficulties in selective updating.

To overcome the limitations of parametric memory, researchers have developed external memory augmentation mechanisms\cite{yang2024text}. These approaches effectively compensate for LLMs' constraints in context length and knowledge integration capabilities by constructing sustainable, precisely accessible external knowledge systems. The development of external memory demonstrates distinct phased characteristics: initially represented by retrieval-augmented generation (RAG)\cite{lewis2020retrieval}, which stores unstructured text fragments in vector databases - while partially addressing knowledge updating issues, it only achieves shallow semantic matching. As research deepens, the focus has gradually shifted from simple information storage to systematic memory management, giving rise to innovative architectures such as GraphRAG's\cite{edge2024local} knowledge graph-based structured memory, MemGPT's\cite{packer2023memgpt} operating system-like memory management, and Zep\cite{rasmussen2025zep} systems that further integrate temporal knowledge engines to support joint retrieval of historical relationships and dynamic contexts, significantly enhancing the reasoning capabilities and management efficiency of memory systems.

Memory mechanism research has now evolved along several key dimensions: in representation and storage, exploring optimal balances between text, summaries, and structured data formats; in indexing and retrieval, developing composite patterns that integrate temporal, semantic, and multi-dimensional features; in update and forgetting mechanisms, investigating intelligent evaluation methods based on information value and timeliness; and in memory utilization and reasoning, focusing on seamless integration between external knowledge and the model's internal reasoning processes. Simultaneously, establishing systematic evaluation frameworks has become crucial for advancing this field.

Despite significant progress in memory augmentation technologies, general architectures face particular challenges when adapting to stock prediction tasks. In memory representation, existing systems struggle to achieve fine-grained modeling of event sequences; in retrieval mechanisms, they lack effective matching capabilities for historical scenario patterns; and in temporal processing, they cannot adapt to the unique dynamic evolution patterns of financial markets. These systematic limitations highlight the urgent need to develop specialized memory architectures specifically designed for the financial domain.

\subsection{LLM in stock prediction}
%%%%！！！！[凝练，介绍方法为主，不足之处去掉intro部分已有的冗余]
This chapter will survey the relevant literature along two primary dimensions: first, the transition from historical pattern matching to memory mechanisms, and second, the advancing paradigm of event-driven prediction. Although existing studies have made remarkable advances, significant limitations remain in integrating structured event knowledge with traceable memory mechanisms - a critical gap that establishes the central contribution of our work.
\subsubsection{From Historical Pattern Matching to Memory Mechanisms}
In the field of time series forecasting, stock price prediction methods based on identifying similar historical patterns had garnered significant academic attention well before the emergence of LLMs. The core concept of these approaches — identifying patterns from historical data that resemble current contexts to inform predictions — is conceptually aligned with the Memory mechanism in LLMs. Both paradigms share the fundamental objective of effectively storing and retrieving external experiences or historical contexts. Early research primarily relied on statistical methods. 
%, using algorithms such as Dynamic Time Warping (DTW) and autoregressive models to identify recurring patterns in price sequences. 
While these methods could effectively capture numerical pattern characteristics, they lacked the capacity to process complex market information.

With the development of deep learning, researchers began employing networks like LSTM and Transformer to automatically learn temporal features. 
%For instance, the DTW-SACP-LSTM\cite{zipingnews} model integrated DTW's sequence alignment capability with LSTM's temporal modeling, enhancing prediction accuracy while maintaining interpretability. 
Transformer-based temporal models effectively captured long-term dependencies through self-attention mechanisms, further advancing the field. The FinSeer framework additionally introduced LLMs to retrieve numerically similar historical sequences, achieving more profound feature mining.

However, these methods exhibit significant limitations in two core dimensions. Regarding data types, most studies still primarily rely on numerical time series, with textual information utilization limited to shallow feature extraction, making it difficult to fully integrate multi-source information such as market sentiment and policy interpretation. In terms of model mechanisms, although neural networks enhance feature learning capabilities, they lack deep comprehension of the semantic logic underlying scenarios, unable to establish causal relationships between event logic and price movements, thus limiting the model's adaptability in complex market environments.

The emergence of LLMs has promoted a paradigm shift from historical pattern matching to memory mechanisms\cite{yu2025finmem, yu2024fincon}. FinMem\cite{yu2025finmem} constructed a specialized financial memory bank, achieving preliminary memory functions by storing news summaries and event embeddings. MagicNet\cite{luo2025magicnet} introduced a hierarchical memory architecture, storing market information in layers according to time granularity. However, these methods still face limitations in memory structure design: memory units mostly consist of isolated text fragments or embedding vectors, lacking complete modeling of continuous event development trajectories; retrieval mechanisms rely on semantic similarity, making it difficult to support complex scenario analogies.%%【结合一下针对本任务的不足，参考intro部分】

Overall, in the evolutionary process from historical patterns to memory mechanisms, existing methods still require deepening in mining and understanding effective information, which provides a clear direction for this paper's research.
%%%Our work distinguishes itself by focusing on the core of what drives price movements: structured events and their sequences. By building a memory of events and their consequences, and by retrieving analogous historical sequences, our framework allows the LLM to perform reasoning that is grounded in concrete event dynamics and market reactions, offering a more interpretable and powerful approach to financial forecasting.
\begin{figure*}[h]
\small
  \centering
  \includegraphics[width=\linewidth]{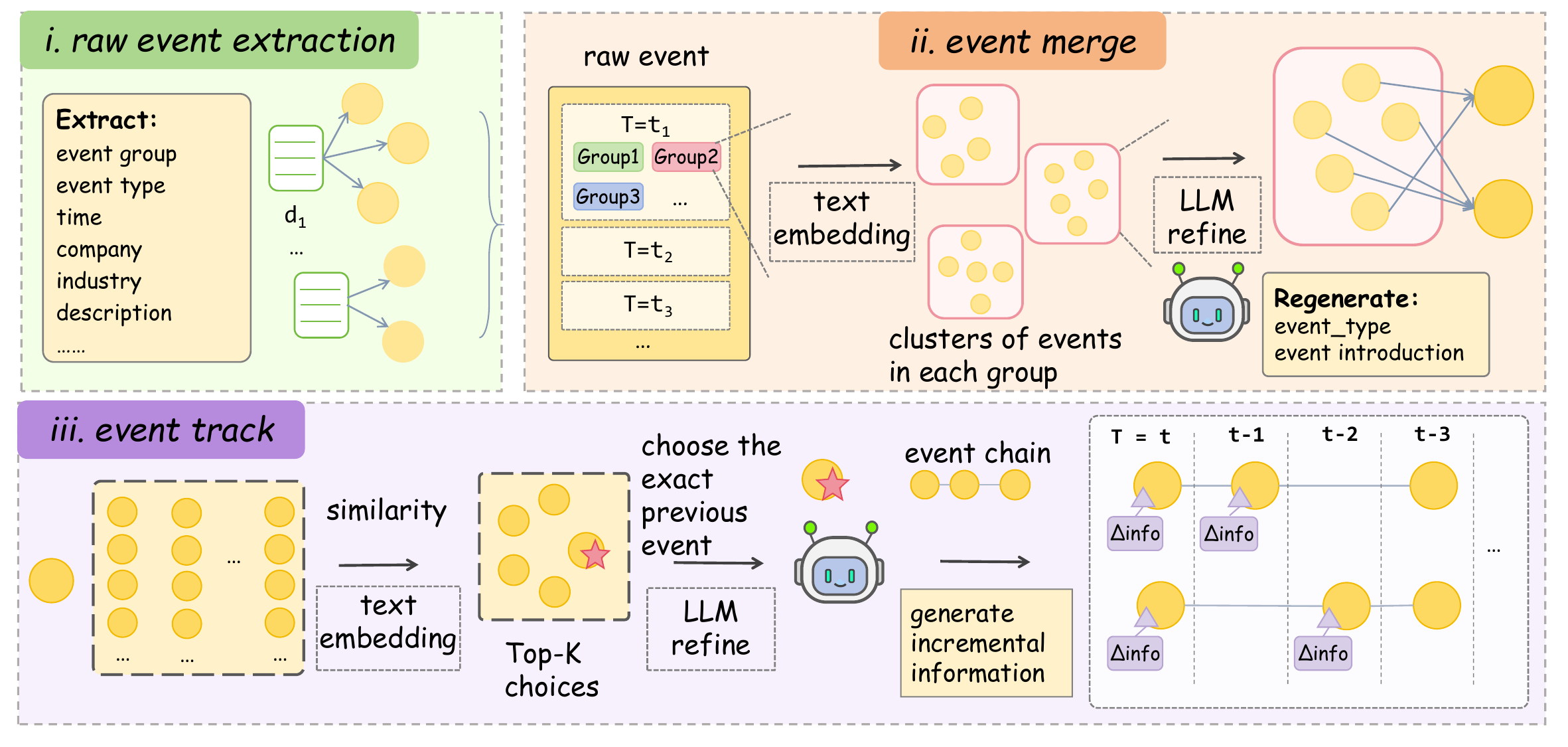}
  \caption{The workflow for constructing the Event Memory, which includes event extraction, merging, and tracking.}
  \Description{event_memory}
  \label{fig:event memory}
\end{figure*}
\subsubsection{Event-Driven Stock Prediction}
The event-driven paradigm aims to identify the fundamental causal mechanisms that drive stock price movements. 
In the pre-LLM era, this field evolved from event study methodology based on cumulative abnormal returns to the automatic identification of event signals from text using Bag-of-Words\cite{zhang2010understanding} models, topic models\cite{blei2009topic} , and eventually deep learning representations like BERT\cite{koroteev2021bert} . 

%The event-driven paradigm represents another crucial research thread, with its core objective being the identification of fundamental causal mechanisms that drive stock price movements. The development of this field clearly demonstrates an evolutionary path from macro-statistical correlations to micro-semantic reasoning.
%In the pre-LLM era, research evolved from statistical validation to feature engineering. Early event study methodology established the statistical significance of the impact of specific corporate events on stock prices through calculating cumulative abnormal returns, providing empirical foundation for the "information drives prices" theory. With advancements in natural language processing, the research focus gradually shifted towards automatically identifying event signals from unstructured text—evolving from bag-of-words\cite{zhang2010understanding} models and topic models\cite{blei2009topic} to deep BERT-based\cite{koroteev2021bert} representations, progressively enhancing text representation capabilities. 
However, a common fundamental limitation persists across these methods: they essentially remain within a "numerically-driven" paradigm\cite{sundermeyer2012lstm,kalyan2021ammus}. Regardless of how event labels or sentiment scores are extracted, they are ultimately fed as features into temporal models such as LSTM or Transformer. Consequently, the models learn statistical correlations rather than event logic, suffer significant loss of critical semantic details during the featureization process, and lack deep reasoning capabilities.

%However, when researchers attempted to input event labels or sentiment scores as features into temporal models like LSTM or Transformer, these methods essentially remained within a "numerically-driven" paradigm\cite{sundermeyer2012lstm,kalyan2021ammus}. Models learned statistical correlations rather than event logic, with critical semantic details being largely lost during the featureization process, and they lacked deep reasoning capabilities.
The advent of LLMs has created opportunities for a paradigm shift in event-driven prediction. Unlike traditional methods, LLMs can extract complete event structures—including elements such as type, subject, and time—directly from raw text. It can preserves information integrity to support causal inference and provides traceable evidence for reasoning. 
%The advent of the LLM era brought technological breakthroughs that created opportunities for paradigm shifts in event-driven prediction. Compared to traditional methods, LLMs can extract complete event structures—including elements such as type, subject, and time—directly from raw text. This capability is revolutionary: it both preserves information integrity to support causal inference and provides traceable reasoning evidence. 
Nevertheless, existing research has yet to fully leverage the potential of LLMs. Approaches represented by CAMEF\cite{zhang2025camef}, while utilizing LLMs for event extraction, subsequently compress the structured events into low-dimensional vectors for processing by traditional neural networks. This effectively reduces LLMs to advanced feature extractors. Such an approach leads to the loss of event details during vectorization, lacks explicit modeling of event relationships and temporal dynamics, and fails to form traceable, reusable experiential memory.
%However, existing research has yet to fully leverage the potential of LLMs. Approaches represented by CAMEF\cite{zhang2025camef}, while utilizing LLMs for event extraction, subsequently compress structured events into low-dimensional vectors for processing by traditional neural networks. This essentially reduces LLMs to advanced feature extractors. Such an approach leads to loss of event details during vectorization, lacks explicit modeling of event relationships and temporal dynamics, and fails to form traceable, reusable experiential memory, ultimately limiting the model's ability to directly learn from historical similar event sequences.
Our work differs fundamentally from the aforementioned approaches. Instead of compressing rich event structures into flattened embedding vectors, we leverage them as the fundamental units for constructing a structured event memory. By explicitly modeling event merging, tracking, and sequencing, our framework enables direct reasoning about event evolution trajectories and analogies across sequences. This addresses the deficiencies of existing LLM-based methods in event sequence correlation analysis and cross-cycle pattern matching, constituting the core contribution of this paper.

\section{Framework}
We introduce a structured event reasoning framework for stock price prediction. 
We dynamically construct an event knowledge base from news corpora and enhance it with model feedback. This allows the framework to retrieve and reason from historically analogous event patterns. The overall architecture is shown in Figure 1 and described in the following subsections.
%Its core mechanism constructs and maintains a dynamic event knowledge base from news, which is continually refined with model feedback. This design empowers the model to perform analogical reasoning by identifying and leveraging historically similar event sequences, moving beyond surface-level text matching. The architecture is outlined in Figure 1 and elaborated upon below.
%This paper proposes a financial stock price prediction framework based on structured event reasoning. The core concept of this framework is to construct a dynamically evolving temporal event knowledge base from unstructured news corpora, incorporating model feedback, and leveraging experiences from historically similar patterns for analogical reasoning. The overall architecture is shown in Figure\ref{fig:framework}, and the following subsections will introduce its core components.
\subsection{Event Memory}
\label{sec:subsection}
Event memory is the core module of our framework. In this section, we transform a continuous stream of unstructured financial news into structured temporal sequences of market events. Figure~\ref{fig:event memory} illustrates the detailed workflow for constructing event memory.

\subsubsection{Event Extraction}
\label{sec:subsubsection}
%\subparagraph{}
We first adopt a technical approach that integrates automatic iterative induction by a LLM with manual correction to evolve a hierarchical event type system from the raw news corpus in the training set
%, following the methodology proposed in [citation]
.
The final system consists of 13 event groups, each further refined into multiple specific event types, amounting to 57 event types in total. 
Detailed definitions are provided in Appendix A.

Each event $E$ is defined as a structured tuple composed of the following elements:

%\begin{itemize}
%\item 
$\bullet$  Required Parameters: event group, event type, time, location, participating entities, involved industries, and involved companies;

%\item 
$\bullet$  Open Parameter: extended attributes dynamically supplemented based on the event context;

%\item 
$\bullet$  Event Description: a summarized text generated from the original news article, intended for text vectorization in subsequent processing steps.
%\end{itemize}

\subsubsection{Event Merging}
\label{sec:subsubsection}
%\paragraph{Paragraph}
%\paragraph{}
%This is a paragraph.
%\subparagraph{Subparagraph}
%This is a subparagraph.
%\subparagraph{}
For each news document $d$, we employ $LLM_{ext}$ to extract events, obtaining a raw event set 
%$E^{\text{raw}}_d = \{E^{\text{raw}}_{d,1}, E^{\text{raw}}_{d,2}, ...\}$. 
%For each news document $d$, we employ $\text{LLM}_{\text{ext}}$ to extract events, obtaining a raw event set 
$E^{\text{raw}}_d = \{E^{\text{raw}}_{d,i}\}_{i=1}^{n_d}$, where $n_d$ denotes the number of events extracted from document $d$.
Given that multiple news articles may report on the same event or describe the same event from different perspectives, the raw event sets exhibit information fragmentation. To achieve information integration, we perform event merging on a daily basis, transcending the limitations of individual news articles and enabling effective fusion of daily information.

For a specific date {T=t}, the daily raw event set is defined as:
%${E^{raw}_t = E^{raw}_{d_1} \cup E^{raw}_{d_2} \cup ... \cup E^{raw}_{d_k}}$.
${E^{\text{raw}}_t = \bigcup_{d \in D_t} E^{\text{raw}}_d}$, 
where $D_t$ denotes the set of all news documents published on date $t$.

%%%%%%%
To address the challenge of merging a large volume of events while preserving the core discriminative capability of the LLM against long-context interference and controlling computational costs, we employ a two-stage strategy of "vector-based coarse clustering + LLM-based fine-grained judgment." The specific procedure is as follows:
%The event merging process comprises the following steps:
%%step1
%Vectorization and Clustering: For each event group $G=g$, we convert event descriptions into vector representations and perform preliminary clustering:

$\bullet$ First, event descriptions are vectorized\cite{chen2024bge}: each structured event description is encoded into an embedding vector $\mathbf{v}(E)$. After grouping the raw events into predefined event groups $G = g$, cluster analysis is performed within each group:
\begin{equation}
\mathcal{C}_{t,g} = \mathrm{Cluster}\left( \{ \mathbf{v}(E) \mid E \in E^{\text{raw}}_{t,g} \} \right)
\end{equation}
where 
%$\mathcal{C}_{t,g} = \{C_{t,g,1}, C_{t,g,2}, \dots, C_{t,g,n_{t,g}}\}$ denotes the set of $n_{t,g}$ clusters obtained for group $g$ at time $t$, and 
$E^{\text{raw}}_{t,g}$ is the set of initial events in group $g$.
%%step2
%Fine-grained Cluster Partitioning: Each preliminary cluster $C_{t,g,i} \in \mathcal{C}_{t,g}$ is further processed by $LLM_{merge}$ to identify and separate distinct events:
%%step3
%$\{E_{t,g,i,1}, E_{t,g,i,2}, \dots\} = LLM_{merge}(C_{t,g,i})$
%Event Description Unification and Type Reassignment: For each merged event $E_{t,g,i,j}$, the LLM generates a unified description and reassigns its appropriate event type.

$\bullet$ Subsequently, $LLM_{merge}$ is applied to perform fine-grained analysis of each cluster, identifying and separating logically distinct events within the cluster. Finally, the LLM generates a unified descriptive text for each merged event and recalibrates its event type categorization.

Through this process, we obtain the refined daily event set $E_t = \{E_{t,1}, E_{t,2}, \dots, E_{t,{n_t}}\}$, where ${n_t}$ represents the total number of merged events for date $t$.

\subsubsection{Event Tracking}
%\subparagraph{}
Events in financial markets exhibit significant temporal continuity, often undergoing multiple days of continuous coverage and evolution after their initial occurrence. 
Consequently, even if an event inherently carries strong positive sentiment, its marginal impact on future stock prices may diminish substantially if it has already been fully absorbed by the market over a preceding period. This implies that stock price movements depend not only on the absolute sentiment polarity (positive/negative) of information but, more critically, on the magnitude of its deviation from existing market expectations (i.e., whether it exceeds or falls short of expectations). 
We define the new developments and changes in a current event relative to its historical counterparts as the event's \textbf{incremental information} ($\Delta Info$).

Existing text-based stock prediction studies, while often utilizing historical information within a time window, suffer from two main limitations: firstly, unstructured text processing leads to the loss of critical event details; secondly, they lack continuous tracking of the complete developmental trajectory of events within the window and fail to explicitly provide this evolutionary information to the prediction model. These constraints prevent existing methods from effectively leveraging incremental information.

%To address these issues, our framework explicitly models the dynamic evolution of events. For any event $E_{t,i}$ on day $t$, the system automatically tracks its chain of homologous events within the historical time window $[t-n, t-1]$. The specific procedure is as follows:
To address the aforementioned issues, the framework proposed in this paper explicitly models the dynamic evolution process of events. We analyze events sequentially according to chronological order for each day. For any event $E_{t,i}$ on day $t$, the system automatically tracks its homologous event chains within the historical time window $[t-w, t-1]$. For the purpose of event merging in the system process, this phase still adopts a two-stage strategy of "vector-based retrieval for coarse screening + LLM for fine judgment," as detailed below:

%%%%%[这里和merge都是用事件的简介部分计算文本向量]
$\bullet$ Candidate Event Retrieval: Retrieve Top-K candidate events via vector similarity:
%$\mathcal{K} = \text{TopK}\left(\text{sim}(\mathbf{v}(E_{t,i}), \mathbf{v}(E_{\text{hist}}))\right), \quad E_{\text{hist}} \in E_{t} \cup E_{t-1} \cup \dots \cup E_{t-w}$
\begin{equation}
    \mathcal{K} = \operatorname{Top-}K_{\substack{E_{\mathrm{hist}} \in E_{[t-w, t-1]}}} \left( \operatorname{sim}(\mathbf{v}(E_{t,i}), \mathbf{v}(E_{\mathrm{hist}})) \right)
\end{equation}
%\begin{equation}
%        \mathcal{K} = \underset{E_{\text{hist}} \in E_{[t-w, t-1]}}{\text{Top-}K} \left( \text{sim}(\mathbf{v}(E_{t,i}), \mathbf{v}(E_{\text{hist}})) \right)
%    \end{equation}
where $E_{[t-w, t-1]} = E_{t-1} \cup E_{t-2} \cup \dots \cup E_{t-w}$ denotes the set of all historical events in the window.

$\bullet$ Event Chain Construction: 
%Using $LLM_{track}$, the candidate set $\mathcal{K}$ is analyzed to identify the direct preceding event $E{\text{prev}}$ of the current event. This process can be executed recursively, tracing back up to a threshold of 5 hops to construct a complete event chain $\mathcal{L}$. All historical events in this chain form the comprehensive past context for $E_{t,i}$.
Based on $LLM_{track}$, semantic discrimination is performed on the candidate set $\mathcal{K}$ to identify whether the current event $E_{t,i}$ has a direct predecessor event $E^{prev}_{t,i}$. If it exists, $E^{prev}_{t,i}$ is added to the event chain $\mathcal{L}_{t,i}$ of $E_{t,i}$. 
Then starting from $E^{prev}_{t,i}$, tracing continues backward recursively, reusing previously constructed chains. 
%and with $E_{\text{prev}}$ as the new starting point, the tracing continues backward by reusing the event chain already constructed for this event in previous analyses. 
To control the complexity of the event context and ensure the model focuses on recent key information, 
a maximum trace depth $D_{\text{max}} = 5$ is enforced, yielding a concise recent evolution chain $\mathcal{L}_{t,i}$.
%a maximum trace depth of 5 hops is set, thereby constructing a complete recent evolution chain $\mathcal{L}$ for $E_{t,i}$.

$\bullet$ Incremental Information Extraction: 
%The model finally analyzes the incremental information of $E_{t,i}$ relative to $\mathcal{L}_{t,i}$:
The model finally analyzes the incremental information $\Delta info_{t,i}$ of the current event $E{t,i}$ relative to its historical context $\mathcal{L}_{t,i}$.

The resulting $\Delta info$ serves as a key metric for quantifying the degree to which information deviates from market expectations, providing a crucial basis for accurate stock price prediction.

\subsection{Reflection Memory}
%%%[再润色一下]
Our ultimate objective is to enable the LLM to autonomously learn a knowledge base for analyzing stock price movements based on current facts, thereby equipping it with the capability to predict stock price changes.
After constructing a detailed event memory bank as situational knowledge, we provide both the true labels and event knowledge to the model for analysis, mining the causal relationships between event sequences and stock price fluctuations. This enables the model to learn how to respond in different market environments. We treat the model's feedback as experiential knowledge, building it into a feedback knowledge bank for future test cases to reference historical experiences.
For each sample in the test set, we only use data preceding it during prediction. After prediction, the sample is analyzed by the model together with its true label and incorporated into the knowledge bank, achieving dynamic expansion of the knowledge base.
To enhance the association between events and stock price movements, we utilize training set data to generate causal explanations. Given an event sequence 
%$E_{[t-w,t]}$ 
within a time window $[t-w,t]$ and the incremental information $\Delta info$ on day t, the model $LLM_{reason}$ combines this with the actual stock price movement $\Delta P_{t+1} $to analyze and output both the reasons for the price change and the key events as reflections.
This process endows historical event sequences with causal relationships, forming retrievable "experiences" that constitute the feedback knowledge bank.

\subsection{Retrieval}
Based on the event-feedback dual-layer memory bank constructed in previous sections—which encapsulates the model's learned knowledge of how event sequences influence stock price movements—the system now possesses a knowledge framework analogous to that of a human professional analyst, providing a reliable knowledge foundation for subsequent predictions. When a new sample is introduced, the system first retrieves analogous patterns from the knowledge base to obtain corresponding analytical experience.
We model event dynamics as a temporally ordered sequence. For a given day $t$, the event series is constructed by aggregating events from the previous $w$ days, forming a chronological list: $
Series_t = [E_{t-w}, E_{t-w+1}, ..., E_t]$

\textbullet { Daily Event Vector}

%$bullet$
To quantify the similarity between event sequences, we encode the event types occurring each day into fixed-length binary vectors, constructed as follows: 
To quantify inter-sequence similarity, we encode daily event occurrences into fixed-length binary vectors. For each day $t$, we construct:

%\begin{itemize}
    %\item 
    A \textbf{type-level vector} $\mathbf{V}_t \in \{0,1\}^M$, where $M$ is the total number of event types. Each dimension $m$ satisfies:
    \begin{equation}
        \mathbf{V}_t[m] = 
        \begin{cases} 
        1, & \text{if at least one event of type $m$ occurs on day $t$} \\
        0, & \text{otherwise}
        \end{cases}
    \end{equation}
    
    %\item 
    A \textbf{group-level vector} $\mathbf{G}_t \in \{0,1\}^G$, where $G$ is the total number of event groups. Each dimension $g$ satisfies:
    \begin{equation}
        \mathbf{G}_t[g] = 
        \begin{cases} 
        1, & \text{if group $g$ has any event on day $t$} \\
        0, & \text{otherwise}
        \end{cases}
    \end{equation}
%\end{itemize}

%For each day $t$, a binary vector $V_t$ of length $M$ is constructed, where $M$ represents the total number of event types in the taxonomy. Each dimension $m$ in the vector corresponds to a specific event type:
%$V_t[m] = 1$, if at least one event of type $m$ occurs on day $t$; otherwise, $V_t[m] = 0$.
%Similarly, a binary vector $G_t$ of length $G$ is generated, where $G$ denotes the total number of event groups. Each dimension $g$ corresponds to a specific event group:
%$G_t[g] = 1$, if at least one event belonging to group $g$ occurs on day $t$; otherwise,$ G_t[g] = 0$.

\textbullet { Similarity Calculation}

We employ a hierarchical strategy that integrates both type-level and group-level information to assess event similarity in a fine-grained manner.

%Single-Day Event Similarity.

The comprehensive event similarity between any two days $t_i$ and $t_j$ is calculated as follows:
%$$TypeSim(i, j) = Jaccard(V_i, V_j)$$
%$$GroupSim(i, j) = Jaccard(G_i, G_j)$$
%$$DailySim(i, j) = \alpha· TypeSim(i, j) + (1 - \alpha) · GroupSim(i, j)$$
\begin{align}
    \mathrm{TypeSim}(t_i, t_j) & = \mathrm{Jaccard}(\mathbf{V}_{t_i}, \mathbf{V}_{t_j}) = \frac{|\mathbf{V}_{t_i} \cap \mathbf{V}_{t_j}|}{|\mathbf{V}_{t_i} \cup \mathbf{V}_{t_j}|} \\
    \mathrm{GroupSim}(t_i, t_j) & = \mathrm{Jaccard}(\mathbf{G}_{t_i}, \mathbf{G}_{t_j}) = \frac{|\mathbf{G}_{t_i} \cap \mathbf{G}_{t_j}|}{|\mathbf{G}_{t_i} \cup \mathbf{G}_{t_j}|} \\
    \mathrm{DailySim}(t_i, t_j) & = \alpha \cdot \mathrm{TypeSim}(t_i, t_j) + (1 - \alpha) \cdot \mathrm{GroupSim}(t_i, t_j)
\end{align}

Here,$ \alpha \in [0,1]$ is a hyperparameter that controls the weighting between precise event-type matching and group-level thematic matching. We set $ \alpha = 0.7 $.

%Event Sequence Similarity.
For two event sequences $Series_{t_a}$ and $Series_{t_b}$, both spanning a window of $w$ days, 
%their similarity is obtained by averaging the daily similarities across aligned days within the window:
%$$SeqSim(Series_A, Series_B) = \frac{1}{L} \sum_{k=0}^{L-1} DailySim(V_{a-k}, V_{b-k})$$
their sequence-level similarity is obtained by averaging daily similarities across temporally aligned positions:
\begin{equation}
    \mathrm{SeqSim}(Series_{t_a}, Series_{t_b}) = \frac{1}{w} \sum_{k=0}^{w-1} \mathrm{DailySim}(t_a-k, t_b-k)
\end{equation}

\textbullet{ Retrieval Process}

Based on the similarity metrics described above, the system executes a two-stage retrieval strategy:
%Building upon the similarity metrics defined above, the system executes a two-stage retrieval strategy to identify the most relevant historical experiences.

Coarse Screening: The system retrieves the top $K$ candidate sequences from the historical knowledge base that exhibit the highest similarity to the current event sequence, denoted as:

\begin{equation}
\begin{split}
&\{\text{Series}_{\text{hist}}^{(1)}, \text{Series}_{\text{hist}}^{(2)}, \dots, \text{Series}_{\text{hist}}^{(K)}\} \\
&\quad = \underset{\text{Series}_{\text{hist}} \in \mathcal{M}}{\text{Top-}K} \left( \mathrm{SeqSim}(\text{Series}_{\text{current}}, \text{Series}_{\text{hist}}) \right)
\end{split}
\end{equation}

where $Mem$ denotes the historical memory bank and $\text{Series}_{\text{current}}$ is the current event sequence.
%$Series_{hist_1}$,..., $Series_{hist_K}$.
%${S_{t{hist1}}, \mathcal{S}{t_{hist2}}, \dots, \mathcal{S}{t_{hist_K}}}$

Fine-Grained Judgment: A dedicated retrieval-augmented large language model $\mathrm{LLM}{\text{retrieve}}$ is employed to perform precise discrimination among the candidate sequences, selecting those with genuine reference value:

%$Series_{ref} =Retrieve(Series_{current},  Series_{hist_1}, ..., Series_{hist_K};  LLM_{retrieve})$ 

A dedicated language model $LLM_{retrieve}$ then filters the candidate set through precise semantic discrimination, ultimately selecting a subset of sequences that serve as valid references for the current event sequence:
\begin{equation}
    Series_{ref} = LLM_{retrieve} \left( Series_{current}, \{ Series_{hist}^{(1)}, \dots, Series_{hist}^{(K)} \} \right)
\end{equation}

%$\mathcal{S}{\text{ref}} = \mathrm{Retrieve}\left(\mathcal{S}{\text{current}}, {\mathcal{S}{\text{hist}1}, \dots, \mathcal{S}{\text{hist}K}}; \mathrm{LLM}{\text{retrieve}}\right)$

%The final reference sequences $S_{ref}$ encompass both the historical events from analogous contexts and their corresponding market feedback, thereby providing a reliable foundation for analyzing the evolution of current events and predicting their market impact.
The final reference sequences $Series_{ref}$ contain historically similar events, with their corresponding market feedback denoted as $Reflection_{ref}$, thereby providing a reliable foundation for analyzing the evolution of current events and predicting their market impact.

\subsection{Inference}

The final prediction is generated by $LLM_{predict}$ through synthesizing the following three types of information:

$\bullet$ Recent Event Sequence $Series_{current}$: Describes the event background that may affect stock prices in the recent period.

$\bullet$ Today's Incremental Information $\Delta info_{current}$: Reflects the degree of information change relative to market expectations.

$\bullet$ Historical Reference Experience $Reflection_{ref}$: Provides precedent market reactions under similar patterns.

The final stock price change prediction $\hat{\Delta P}_{t+1}$ is given by:
\begin{equation}
\hat{\Delta P}_{t+1} = LLM_{predict}(Series_{current}, \Delta info_{current}, Reflection_{ref})
\end{equation}

\section{Experiment}
We evaluate the performance the proposed framework and delve into its internal mechanisms through a series of experiments.  Our investigation primarily revolves around the following questions:

$\bullet$ RQ1: How does the overall performance of our framework compare with existing state-of-the-art memory-augmented models on financial stock prediction tasks?

$\bullet$ RQ2: Does the structured event representation employed in our framework offer advantages over traditional representations such as text summaries or topic clustering?

$\bullet$ RQ3: Is the incremental information obtained through longitudinal event tracking a key factor in improving prediction performance?

$\bullet$ RQ4: In retrieving historical experiences, is the strategy of referencing similar event sequences across companies more effective than the strategy limited to a single company's history?
\subsection{Experimental Settings}
\subsubsection{Dataset}
This study constructs an experimental dataset based on the CSMAR China Stock Market Research Database, which systematically collects daily online news from mainstream financial media.
%and provides a standardized news event classification system.
%We selected iFlytek (002230) and other leading technology stocks as research subjects, primarily based on the following considerations: (1) These stocks exhibit high market attention and liquidity, capable of reflecting the overall trend of the technology sector; (2) Their stock prices are highly sensitive to news events such as industry policies and technological innovations, which is conducive to validating the effectiveness of event-driven prediction; (3) As representative enterprises in their respective fields, they benefit from high news coverage and data quality.
%The experimental time range is set from the first quarter to the second quarter of 2024, with 
%The training period spanning from January 1, 2024, to March 31, 2024, and the test period from April 1, 2024, to June 30, 2024. In the data processing pipeline, we strictly adhere to the chronological order of the time series: for each sample in the test set, after completing the daily prediction, both the sample and its true label are added to the training set. This enables the model to acquire market feedback through an online learning mechanism, simulating the continuous learning process in real investment environments.
We employed a rolling-window evaluation strategy, where the model was trained on data from the first quarter of 2024 (January 1 to March 31) and tested on data from the second quarter (April 1 to June 30). To simulate real-time learning, after each test instance was predicted, both the instance and its true label were provided to the model for training and incrementally added to the memory, thereby establishing an online learning feedback loop.
We conducted experiments on four technology companies: iFlytek, Industrial Fulian, Cambricon, and Hygon. Daily returns were used to measure price movements, with a ±1\% threshold applied to classify directions due to the sector's high volatility. Returns above 1\% were labeled as "up", below -1\% as "down", and those within [-1\%, 1\%] as "flat" which were excluded during testing.

\subsubsection{Models}
We employs DeepSeek-V3\cite{liu2024deepseek} as the core language model, while text vectorization utilizes the BGE-M3\cite{chen2024bge} model.
\subsubsection{Baselines}
To comprehensively evaluate the performance of the proposed framework, we selected five representative memory-augmented models as baselines:

$\bullet$ FinMem\cite{yu2025finmem}: %A memory system specifically designed for the financial domain, implementing memory functions through storing news summaries
A financial-domain memory system that implements a layered memory structure (short/mid/long term) to guide trading decisions

$\bullet$ MemP\cite{fang2025memp}: 
A general framework that abstracts trajectories into both step-by-step instructions and high-level strategies, thereby maintaining and utilizing procedural memory.
%A general memory model based on parametric memory architecture

$\bullet$ MemGPT\cite{packer2023memgpt}: 
An OS-inspired memory system that manages hierarchical storage and interrupt handling to dynamically expand the LLM's effective context window.
%A memory system that adopts operating system-inspired memory management, providing the LLM with an expandable context window through hierarchical memory and interrupt handling.
%A memory system adopting operating system memory management concepts

$\bullet$ Zep\cite{rasmussen2025zep}: 
A long-term memory framework incorporating temporal knowledge graphs to support joint retrieval of historical relationships and dynamic contexts
%A long-term memory framework incorporating temporal knowledge graphs
\subsubsection{Evaluation Metrics}
Accuracy (ACC) and Matthews Correlation Coefficient (MCC) are adopted as evaluation metrics. ACC measures the overall prediction correctness, while MCC serves as a reliable metric for balanced datasets, providing more robust performance assessment in stock prediction tasks where class distribution may be imbalanced.

\subsection{Overall Performance Comparison (RQ1)}
%4.2 Overall Performance Comparison (RQ1)
We conducted evaluations on four representative stocks. As shown in Table~\ref{tab:performance_comparison} , our framework consistently and significantly outperformed all baseline models in ACC and MCC,
which validates the overall effectiveness of our approach.
%These results fully validate the overall effectiveness of our framework. 

The results further reveal fundamental mismatches between baseline architectures and stock prediction requirements. Specifically, FinMem's isolated news storage lacks temporal event modeling, causing its retrieval mechanism to miss critical incremental information by failing to match evolving market contexts. MemP's procedural abstraction discards semantic event details during trajectory compression, undermining financial scenario reasoning. MemGPT's general-purpose memory management lacks event-aware retrieval, frequently overlooking financially significant patterns during memory updates. Zep's graph-based approach fails to capture inter-event dependencies, missing decisive correlations in financial environments.

Our framework addresses these limitations through a 
%transformation-and-reasoning 
new 
paradigm that converts unstructured news into structured temporal event knowledge. This approach systematically captures event logic and dependencies, establishing a traceable, reasoning-capable system that mirrors human analytical processes for more accurate and trustworthy financial predictions.
%Our framework addresses these limitations through transforming unstructured news streams into structured temporal event knowledge bases, and performing memory retrieval and reasoning based on this foundation，witch proves to be a superior approach for financial information processing compared to traditional or general-purpose memory methods. This demonstrates that through in-depth event-based information analysis, constructing a traceable and reasoning-capable professional knowledge system analogous to human analysts is crucial for achieving more accurate and trustworthy financial predictions.

%\begin{table}
%  \caption{Frequency of Special Characters}
%  \label{tab:freq}
%  \begin{tabular}{ccl}
%    \toprule
%    Non-English or Math&Frequency&Comments\\
%    \midrule
%    \O & 1 in 1,000& For Swedish names\\
%    $\pi$ & 1 in 5& Common in math\\
%    \$ & 4 in 5 & Used in business\\
%    $\Psi^2_1$ & 1 in 40,000& Unexplained usage\\
%  \bottomrule
%\end{tabular}
%\end{table}

\begin{table*}[h]
\centering
\small
\caption{Performance comparison (ACC and MCC) between StockMem and baseline models on four stocks.}
\label{tab:performance_comparison}
%\begin{tabular*}{\textwidth}{@{\extracolsep{\fill}}lcccccccc}
\begin{tabular}{lcccccccc} 
\toprule
\multirow{2}{*}{Model} & \multicolumn{2}{c}{iFlytek} & \multicolumn{2}{c}{Industrial Fulian} & \multicolumn{2}{c}{Cambricon} & \multicolumn{2}{c}{Hygon} \\
\cmidrule(lr){2-3} \cmidrule(lr){4-5} \cmidrule(lr){6-7} \cmidrule(lr){8-9}
 & ACC & MCC & ACC & MCC & ACC & MCC & ACC & MCC \\
\midrule
Finmem & 0.5763 & 0.1430 & 0.5247 & 0.1056 & 0.5339 & 0.1142 & 0.5129 & 0.1052 \\
MemGPT & 0.5164 & 0.0584 & 0.4973 & 0.0177 & 0.5216 & 0.1002 & 0.5228 & 0.1068 \\
Memp & 0.5271 & 0.0962 & 0.5164 & 0.0782 & 0.5032 & 0.0505 & 0.4897 & 0.0071 \\
Zep & 0.5624 & 0.1298 & 0.5481 & 0.1123 & 0.5181 & 0.0773 & \textbf{0.5493} & \textbf{0.1190} \\
Ours & \textbf{0.6253} & \textbf{0.1692} & \textbf{0.5714} & \textbf{0.1277} & \textbf{0.5622} & \textbf{0.1363} & 0.5403& 0.1186 \\
\bottomrule
%\end{tabular*}
\end{tabular}
\end{table*}

\subsection{Ablation Study}
\subsubsection{Effectiveness of Event Representation Forms(RQ2)}
%To validate the necessity of using structured events as information units (RQ2), we 
We systematically compared three information representations: our structured events, text summaries of original news, and clustered opinions derived from these summaries. As shown in Table ~\ref{tab:ablation_events}, replacing structured events with either alternative led to significant performance degradation, underscoring the critical role of structured event representation in our framework.
%We systematically compared three information representation methods: the structured events utilized by our framework, text summaries generated from original news articles, and clustered opinions formed by further clustering these summaries. The experimental results Table ~\ref{tab:ablation_events} clearly demonstrate that replacing structured events with either text summaries or clustered opinions leads to a significant decline in model performance, strongly confirming the central importance of the structured event representation in our framework.

\begin{table}[h]
\centering
\small
\caption{Ablation Study on Event Representation Forms}
\label{tab:ablation_events}
\begin{tabular}{lcc}
\toprule
Method & Avg\_acc & Avg\_mcc \\
\midrule
Summary & 0.4378 & 0.0012 \\
Cluster opinion & 0.4655 & 0.0157 \\
Event (Ours) & \textbf{0.5748} & \textbf{0.1380} \\
\bottomrule
\end{tabular}
\end{table}
The advantage of structured events lies in their superior information organization and retention:

$\bullet$ Coherent Knowledge Integration: By merging related events and tracing their precedents, structured events form denoised, logically connected event chains that consolidate dispersed information flows. In contrast, summary-based representations remain fragmented even after clustering, lacking causal or temporal linkages.

$\bullet$ Fine-Grained Semantic Preservation: Structured events retain discriminative details during abstraction (e.g., technical parameters in a "new technology release"), whereas summaries generalize and clustered opinions oversimplify such information into ambiguous labels (e.g., "technology release"), erasing features critical for assessing commercial impact.

$\bullet$ Precise Temporal Anchoring: Day-level timestamps in structured events enable precise analysis of lag effects and impact duration. Unstructured summaries often lack clear temporal cues, hindering the establishment of causal relationships between events and market reactions and thus compromising prediction accuracy and interpretability.

\subsubsection{The Role of Event Tracking and Incremental Information (RQ3)}

%This experiment aims to examine the importance of capturing incremental information in event evolution (RQ3). 
We compared the complete framework with an ablated version that removed the event tracking and incremental information module.

The significant performance decline shown in Table ~\ref{tab:ablation_tracking} clearly demonstrates that longitudinal event tracking and the resulting incremental information are indispensable components of the framework, providing a positive answer to RQ3. 
This result strongly aligns with the "expectation discrepancy" theory in finance: price movements depend less on information's absolute nature than on its deviation from market expectations (i.e., $\Delta Info$). Removing event tracking essentially blinds the model to subtle sentiment shifts driven by expectation revisions.
%This finding strongly aligns with the "expectation discrepancy" theory in financial markets - stock price movements are determined not only by the absolute nature of information but more importantly by how information deviates from existing market expectations (i.e., incremental information). Removing this module essentially deprives the model of its ability to perceive subtle shifts in market sentiment.

\begin{table}[h]
\centering
\small
\caption{Ablation Study on Event Tracking and Incremental Information}
\label{tab:ablation_tracking}
\begin{tabular}{lcc}
\toprule
Method & Avg\_acc & Avg\_mcc \\
\midrule
w/o $\Delta$info & 0.4058 & 0.0067 \\
with $\Delta$info (Ours) & \textbf{0.5748} & \textbf{0.1380} \\
\bottomrule
\end{tabular}
\end{table}
Case analysis further illustrates this mechanism:
for instance, a modestly received new service can trigger strong positive returns if subsequent reports show unexpected application progress.
Conversely, repeated news of a major shareholder's holdings reduction—with no substantial developments—elicits neutral market reactions, as no new expectation deviation occurs.
%Case analysis further reveals its value: for instance, when a company initially launches a new service with modest market reaction, subsequent reports indicating significant progress in the service's application can trigger strong positive market responses. Conversely, when news reports about a major shareholder reducing their holdings are repeated over subsequent days without substantial new developments compared to the initial report, the impact on stock price remains neutral.
Thus, while individual events may have limited impact, the incremental shift relative to market expectations—captured through event tracking—holds superior predictive power.
%This demonstrates that while the impact of individual events is often limited, the trend changes identified through event tracking and the resulting "above-expectation" degree possess stronger predictive value.

\subsubsection{Impact of Historical Sequence Matching Strategies (RQ4)}

To evaluate the effectiveness of our historical pattern referencing strategy (RQ4), we compared four different historical experience referencing approaches. The comparative results Table~\ref{tab:ablation_matching} clearly demonstrate the superior performance of our framework's strategy.

\begin{table}[htbp]
\centering
\small
\caption{Ablation Study on Historical Sequence Matching Strategies}
\label{tab:ablation_matching}
\begin{tabular}{lcc}
\toprule
Method & Avg\_acc & Avg\_mcc \\
\midrule
w/o reference & 0.4509 & 0.0089 \\
Recent-Period reference & 0.4077 & 0.0064 \\
Same-company matching only & 0.5686 & 0.1358 \\
Full reference (Ours) & \textbf{0.5748} & \textbf{0.1380} \\
\bottomrule
\end{tabular}
\end{table}

Detailed analysis reveals the following insights: The "same-company matching only" strategy, while achieving reasonable results, suffers from data sparsity limitations. This is particularly evident for companies with shorter histories or innovative business models where finding high-quality similar sequences becomes challenging. Additionally, this strategy exhibits pattern singularity, unable to learn from broader industry experiences.

The "recent-period reference" strategy performed poorly as it ignores the essence of events, relying solely on temporal proximity for reference, making it highly susceptible to interference from recent market random fluctuations.

The "no reference" strategy yielded the worst performance, confirming the fundamental importance of historical experience referencing for this task.

The success of our strategy lies in its ability to enrich reference patterns and improve data utilization. It enables the model to learn more universal patterns from other companies' historical responses when dealing with industry-common events. A representative case demonstrates this: when iFlytek achieved phased progress in "large language model research", our framework successfully referenced the market response patterns experienced by Cambricon during its early technological breakthroughs. Consequently, it more accurately predicted the short-term stock price trends and correction timing compared to models relying solely on iFlytek's limited historical data. This proves that within specific market sectors, predictable and transferable regularities exist in market responses to similar events, which our framework effectively captures and leverages.

\section{Conclusion}

We present StockMem, an event-reflection memory framework for explainable stock forecasting. Our approach overcomes limitations of generic memory architectures by constructing temporal knowledge bases from web news through structured event modeling. The dual-layer design features: (1) Event Memory capturing market event evolution via horizontal consolidation and longitudinal tracking, extracting expectation-discrepancy signals; (2) Reflection Memory abstracting causal experiences from event-price interactions.

Experiments show StockMem significantly outperforms state-of-the-art baselines. Ablation studies confirm the necessity of structured events, longitudinal tracking, and cross-company matching. The framework provides inherent explainability by tracing event chains and market logic, enhancing decision transparency.

This work advances LLM applications in finance and introduces a new paradigm for domain-specific memory systems. Future work will extend to multi-modal financial data and other web-based prediction tasks.

%\begin{acks}
%To Robert, for the bagels and explaining CMYK and color spaces.
%\end{acks}

%%
%%！！！参考文献
\bibliographystyle{ACM-Reference-Format}
%\bibliography{sample-base}
\bibliography{reference}

\appendix

\section{Event type}
To construct a precise event taxonomy suitable for the financial domain, this study must overcome the dual challenges of domain expertise and taxonomic completeness. Directly adopting external classification systems may not align with the specific characteristics of our dataset, while a fully manual construction approach is both cost-prohibitive and prone to subjectivity. To address this, we designed a semi-automated methodology termed "LLM-driven Iterative Induction and Human Correction." This approach leverages the semantic induction and classification capabilities of large language models to automatically evolve a hierarchical event taxonomy from raw news corpora.

In each iteration, using the current taxonomy as a baseline, the LLM processes all news articles in batches to perform two core tasks: 1) Judge whether the event described in the news can be categorized into existing types. 2) Identify and propose new event types not yet covered. The taxonomy is considered stable once the model consistently proposes no new types over multiple consecutive iterations. Subsequently, the research team manually corrects and consolidates the induced results, ultimately yielding a stable taxonomy comprising several event groups and types for subsequent research.

The specific event groups and types are as follows:

$\bullet$ Policies and Regulation: Policy Release, Development Planning, Government Support, Institutional Supervision, International Controls and Sanctions

$\bullet$ Macroeconomic Finance: Fiscal Policy, Livelihood and Welfare, Taxation

$\bullet$ Industry Standards: Standards, Specifications, Opinions and Commentary

$\bullet$ Products and Market: Research and Development, New Product Launch, Product Mass Production, Product Application, Product Price Changes, Product Output Changes, Supply-Demand Dynamics

$\bullet$ Technology Events: Technological Breakthrough, Research and Development Progress, Certification, Shipment, Ecosystem Collaboration, Enablement

$\bullet$ Corporate Operations: Investment, Financing, Expenditure, Profitability, Order Service Agreement Signing, Order Service Agreement Changes, Contracts, Mergers and Acquisitions

$\bullet$ Corporate Projects: Project Initiation, Project Implementation, Cross-sector Expansion

$\bullet$ Corporate Equity: Shareholder Changes, Share Increase, Share Decrease, Ownership Disputes

$\bullet$ Corporate Personnel: Executives, Personnel Changes, Violations and Misconduct

$\bullet$ Stock Market Performance: Market Size, Sector Concept Performance, Individual Stock Performance, Capital Flows, Trading Activities, Institutional Views

$\bullet$ Other Financial Market Performance: Market Size, Market Performance, Capital Flows, Institutional Views

$\bullet$ Cooperation and Strategy: Strategic Cooperation and Co - construction, Industry Alliances and Standards Organizations

$\bullet$ Risks and Warnings: Business Clarification, Company-Specific Risks, Industry-Wide Risk Alerts

\section{Prompt}
$\bullet$ The prompt of train process:

You are a stock analyst specializing in \{stock\}. You need to interpret the driving factors behind tomorrow's stock price movement based on the following analytical elements.

Analytical Elements: Recent event sequence and today's incremental information.
Logic of the Analytical Elements:

The recent event sequence outlines events within a recent time window that may impact tomorrow's stock price.

Incremental information refers to new developments or changes in an event compared to its past occurrences, indicating whether it has become more positive/negative/neutral.

Stock price movements depend not only on the absolute nature of the information (positive/negative) but also on the degree of deviation from existing market expectations (exceeding expectations/falling short of expectations). Incremental information reflects this deviation from market expectations.

=== Events and Incremental Information ===
\{information\}

=== Predicted Direction of Tomorrow's Stock Price Change ===
\{price\_change\}

Please analyze the basis for the stock price change based on the given events and incremental information (within 500 words) and specify which events contributed to the price change (within 300 words).

Output strictly in the following JSON format:
{{"Reason for price movement": xxx, "Events causing the impact": xxx}}

$\bullet$ The prompt of test process:

You are a stock analyst specializing in \{stock\}. You need to predict tomorrow's stock price movement (up/down) based on the following analytical elements.

Analytical Elements: Recent event sequence, today's incremental information, and relevant historical reference experience.

Logic of the Analytical Elements:

The recent event sequence outlines events within a recent time window that may impact tomorrow's stock price.

Incremental information refers to new developments or changes in an event compared to its past occurrences, indicating whether it has become more positive/negative/neutral.

Stock price movements depend not only on the absolute nature of the information (positive/negative) but also on the degree of deviation from existing market expectations (exceeding expectations/falling short of expectations). Incremental information reflects this deviation from market expectations.

Historical reference experience includes similar event sequence patterns matched from historical data based on the characteristics of the current event sequence. These historical similar sequences contain events and incremental information from that period, along with their corresponding subsequent stock price movements, reflecting how the market typically reacts to various types of information in similar situations.

=== Events and Incremental Information ===
\{information\}

=== Historical Reference Experience ===
\{hist\_reflection\}

Please refer to the historical experience and predict the stock price change based on the given events and incremental information. Analyze the basis for the price movement (within 500 words). 

Output strictly in the following JSON format:
{{"Reason for price movement": xxx, "Price movement": up/down}}

\end{document}